\documentclass{article}

\PassOptionsToPackage{round}{natbib}   
\usepackage[dblblindworkshop, final]{neurips_2026}
\workshoptitle{AI for Stochastic Dynamics}

\usepackage[utf8]{inputenc}            
\usepackage[T1]{fontenc}               
\usepackage{amsmath,amssymb,mathtools} 
\usepackage{hyperref}                  
\usepackage{url}                       
\usepackage{booktabs}                  
\usepackage{nicefrac}                  
\usepackage{microtype}                 
\usepackage{xcolor}                    
\usepackage{graphicx}

\newcommand{\MASK}{\texttt{[MASK]}}
\newcommand{\ci}[2]{{\scriptsize$[#1,#2]$}}

\title{Trust but Verify: Designing Region-Inhomogeneous\\ Forward Dynamics for Discrete Diffusion}

\author{%
  Gunmay Jhingran\\
  Delhi Technological University\\
  \texttt{gunmayjhingran\_23ep040@dtu.ac.in}\\
}

\begin{document}

\maketitle

\begin{abstract}
In continuous diffusion, the forward process is a design surface: noise schedules, SDE variants, and per-position noise levels are all engineered to match the task. Discrete diffusion for language has collapsed this surface to a single point, the absorbing process, under which conditioning context sits at infinite signal-to-noise ratio for the entire reverse trajectory. The collapse has a cost. Trust in the conditioning becomes binary, so corrupted context tokens are copied as ground truth and their errors propagate into the generated spans. We treat the discrete forward process as a design space again, composing absorbing and uniform-mixing corruption kernels with region-dependent rates: hole positions traverse the standard absorbing dynamics, while user-supplied context is corrupted by light random substitution at a small rate $\eta$, so that ``observed but uncertain'' is a state the model has visited during training. At inference, a heuristic reverse-time sampler re-injects noise into suspicious context tokens under an annealed intensity (a corrector-like step that acts on the conditioning rather than the sample) and freezes the context dynamics late in the trajectory. An 8{,}000-step finetune of a 170M-parameter pretrained masked diffusion model, under five hours on a 6\,GB laptop GPU, converts a blind copier into a verifier. The base model assigns roughly 50\% probability to observed tokens even when they are uniform-random garbage; on conspicuous (uniform-random) corruptions, the finetuned model detects corrupted context with 0.96 recall at 0.91 precision, improves infill token-F1 by 3.8 points at 10\% context corruption (6.4 at 20\%), and eliminates the accuracy dip in tokens generated adjacent to corruptions, with no measurable regression on clean conditioning (false-edit rate under 0.4\% on clean context). Gains under plausible corruptions are much weaker (about 0.5 F1 points at 10\% corruption, not statistically significant). The experiments also suggest a regularity: detectability and damage are jointly governed by how far the corruption kernel departs from the data conditional. Noise near the conditional is both statistically invisible and semantically light; noise far from it is conspicuous and destructive. We interpret absorbing masked diffusion as the infinite-contrast limit of a trust axis whose continuous analog is the SNR gap between known and unknown regions in diffusion inpainting; in our setting (one 170M model, one text domain), points inside that axis can be reached by perturbing the forward dynamics rather than the architecture.
\end{abstract}

\section{Introduction}
\label{sec:intro}

Continuous diffusion treats its forward stochastic dynamics as an engineering surface. The choice of SDE and noise schedule \citep{sohl2015deep, ho2020denoising, song2021scorebased}, the renoising of observed regions in inpainting \citep{lugmayr2022repaint}, and per-position noise levels as first-class objects \citep{chen2024diffusion} are all instances of the same principle: the corruption process is designed to match the task, and the reverse-time dynamics \citep{anderson1982reverse} inherit its structure. Discrete diffusion began with the same breadth: \citet{austin2021structured} studied whole families of corruption kernels, and continuous-time formulations characterize the general Markov-process setting \citep{campbell2022continuous, lou2024discrete}. In practice, though, the field has collapsed onto a single point in that space, the absorbing process, in which each token independently jumps to \MASK{} and stays there. The reasons are good ones. The absorbing kernel admits a simple weighted cross-entropy bound \citep{sahoo2024simple, shi2024simplified}, its optimal denoiser is time-free, modeling the conditionals of clean data directly \citep{ou2025your}, and any partially observed sequence is in-distribution at every noise level, so conditioning requires no mechanism at all. The recipe scales to LLaDA-class models competitive with strong autoregressive baselines \citep{nie2025llada}.

The collapsed design space has a structural cost that is easiest to see in the language of noise. Under absorbing dynamics, an observed token has, in effect, infinite signal-to-noise ratio: the reverse process carries it with probability one, forever. Trust in the conditioning is therefore binary: a position is either fully certain or fully unknown, and there is no state expressing that context is \emph{mostly} right. Three consequences follow. When the conditioning text contains errors (a corrupted retrieval, a buggy code file, a noisy transcription), the model conditions on wrong tokens as if they were ground truth and the errors contaminate the spans it generates; we find the pretrained model assigns roughly 50\% probability to observed context tokens \emph{even when they are uniform-random garbage}. The model has no mechanism to repair its context, because the reverse dynamics never revisit unmasked states. And graded uncertainty about the condition, which continuous inpainting can express through the elapsed noise of the known region, has no discrete counterpart.

We propose to reopen the design space with the minimal move: compose two corruption kernels with region-dependent rates. Hole positions traverse the standard absorbing dynamics indexed by $t$; user-supplied context is corrupted by the uniform-mixing kernel at a small per-sequence rate $\eta$; equivalently, the context region is held at a fixed, small elapsed time under uniform-mixing dynamics while holes traverse the absorbing process (as in standard masked diffusion, only the forward marginals enter the objective). A short finetune of a pretrained masked diffusion model under this region-inhomogeneous forward process, with an $\eta{=}0$ mixture component protecting the clean-conditioning regime, yields a model that emits a meaningful predictive distribution at \emph{every} position. That distribution supplies a free suspicion score for context, the model's own probability of the observed token, which our sampler exploits through a corrector-like heuristic acting on the conditioning: early in the reverse trajectory, context tokens the model finds implausible are re-noised into the absorbing state under an annealed intensity and refilled by the ordinary reverse dynamics with full bidirectional conditioning; past a lock time, the context dynamics freeze.

We make three contributions. We formalize region-inhomogeneous forward processes for masked diffusion and show they require only a light perturbation of the standard SFT-style objective (Section~\ref{sec:method}). We introduce the detect--remask--refill sampler with trust annealing, a heuristic re-injection of noise into the conditioning during the reverse process, with the confidence bookkeeping needed to compose with low-confidence decoding (Section~\ref{sec:inference}). And we evaluate on corrupted-context infilling over web text with pre-registered hypotheses and decision gates, disentangling training-time from inference-time contributions across nine arms (Section~\ref{sec:experiments}). The evidence is strongest for conspicuous (uniform) corruptions, where the finetune yields detection, repair, and containment of context corruption with no measurable regression under clean conditioning. Under plausible corruptions detection is partial, infill gains are small, and the pre-registered primary F1 gate fails. Our ablations also do not fully separate the contributions of the $\eta{=}0$ mixture and of trust annealing from other factors that change alongside them (Section~\ref{sec:results}). The experiments further suggest a regularity, which we offer as an interpretation rather than a tested result: detectability and damage both appear to be governed by the divergence between the corruption kernel and the data conditional. Conceptually, we view absorbing masked diffusion as one end of a trust axis whose continuous analog is the signal-to-noise contrast between known and unknown regions in diffusion inpainting; this framing motivates the method but is not established formally here.

\section{Related work}
\label{sec:related}

\paragraph{Discrete diffusion as stochastic dynamics.} \citet{austin2021structured} introduced discrete diffusion with structured corruption kernels (absorbing, uniform-mixing, and discretized-Gaussian transition matrices), and continuous-time treatments formalize the general setting as Markov jump processes with learnable reverse rates \citep{campbell2022continuous, lou2024discrete}. Subsequent work specialized to the absorbing kernel, simplifying its variational bound to a weighted masked cross-entropy \citep{sahoo2024simple, shi2024simplified} and establishing the time-free property of its optimal denoiser \citep{ou2025your}; LLaDA scaled this recipe to 8B parameters \citep{nie2025llada}, and block diffusion interpolates between autoregressive and diffusion decoding with clean context to the left of the active block \citep{arriola2025block}. All of these inherit the absorbing process's binary conditioning interface, which is the object of our study.

\paragraph{Hybrid and self-correcting forward processes.} Closest to our forward-process design, GIDD generalizes masked diffusion to arbitrary interpolating noise and trains on a spatially \emph{homogeneous} hybrid of masking and uniform noise, showing the resulting models can identify and correct their own mistakes in generated samples \citep{vonrutte2025generalized}; SCDD reformulates such pretraining-based self-correction with explicit discrete-time transitions \citep{wang2026selfcorrection}; and ReMDM adds inference-time re-noising of \emph{model-committed} tokens to restore iterative refinement \citep{wang2025remasking}. Our setting differs on two axes. The corruption is region-\emph{inhomogeneous}: hybrid noise is applied only to user-supplied context, targeting trust in the \emph{conditioning} rather than the model's own outputs. And the asymmetry matters for reliability: GIDD reports that homogeneous hybrid-noise models lag mask-only models downstream, whereas confining the uniform-mixing kernel to the context region, together with an $\eta{=}0$ mixture, kept clean-conditioning behavior and the held-out variational bound within our pre-registered H3 tolerances at this scale; we did not train a homogeneous hybrid-noise arm, so this contrast is indirect. Our arm B further suggests that, in our setting, inference-time re-noising alone does not create the capability: a model trained only under absorbing dynamics assigns no meaningful likelihood to observed positions.

\paragraph{Denoising and editing text.} SUNDAE trains non-autoregressive generators by unrolled denoising of randomly corrupted sequences \citep{savinov2022step}; DiffusER frames generation as edit sequences \citep{reid2023diffuser}; ELECTRA trains a discriminative replaced-token detector \citep{clark2020electra}. We retain the diffusion formulation: detection falls out of the generative model's own likelihoods rather than a separate discriminator, and repair is performed by the same reverse dynamics that fill holes. Our external-detector baseline (Section~\ref{sec:ext}) gives one point of comparison: a GPT-2 left-context surprisal detector trails the model on conspicuous corruptions and fails on plausible ones. We did not evaluate bidirectional or trained replaced-token detectors such as ELECTRA.

\paragraph{Conditioning in continuous diffusion.} Gaussian inpainting must reconcile clean observations with reverse-time dynamics that expect globally noised inputs, via renoised composition \citep{lugmayr2022repaint} or specialized conditional models; confidence-ordered parallel decoding originates with MaskGIT \citep{chang2022maskgit}; and per-position noise levels as first-class objects appear in Diffusion Forcing \citep{chen2024diffusion}. We view our method as a discrete analog on a shared axis: where those approaches lower the known region from full noise to intermediate SNR, we raise it from infinite SNR to finite.

\section{Method}
\label{sec:method}

\subsection{Background: absorbing forward dynamics}

Let $x_0 \in \mathcal{V}^L$ and let $t \in [0,1]$ index the absorbing forward process: each token independently jumps to the absorbing state \MASK{} so that the forward marginal masks position $i$ with probability $t$. As in standard masked diffusion training, only these marginals enter the objective. The denoiser $p_\theta(\cdot \mid \tilde{x})$ is a bidirectional Transformer trained with the weighted masked cross-entropy of \citet{shi2024simplified, sahoo2024simple}, which upper-bounds negative log-likelihood; by the time-free property \citep{ou2025your}, its optimum is $p_{\mathrm{data}}(x_0^i \mid \tilde{x}^{\mathrm{UM}})$, the conditional of clean data given the unmasked set, independent of $t$. Conditional infilling fixes a set of positions as clean context and runs the reverse process over the remainder; under absorbing dynamics, observed tokens persist with probability one, and that property, its cost and its repair, is our subject.

\subsection{Region-inhomogeneous forward dynamics}

We partition each length-$L{=}512$ sequence into holes $H$ and known context $K$: one to four contiguous spans covering 10--50\% of positions form $H$, matching the infilling use case; the rest is $K$. The two regions evolve under different corruption kernels. Draw $t \sim U(0,1]$ and
\begin{equation}
\eta \sim \tfrac{1}{2}\,\delta_0 + \tfrac{1}{2}\, U(0, \eta_{\max}], \qquad \eta_{\max} = 0.15,
\end{equation}
and corrupt $x_0 \mapsto \tilde{x}$ position-wise:
\begin{equation}
\tilde{x}^i =
\begin{cases}
\MASK{} \text{ w.p. } t, \;\; x_0^i \text{ otherwise}, & i \in H \quad \text{(absorbing kernel)},\\[2pt]
z \sim q_{\mathrm{noise}} \text{ w.p. } \eta, \;\; x_0^i \text{ otherwise}, & i \in K \quad \text{(mixing kernel, } z \neq x_0^i\text{)}.
\end{cases}
\label{eq:forward}
\end{equation}
Holes traverse the absorbing dynamics indexed by $t$; the context region is held at a fixed, small elapsed time under a mixing process, its marginal substitution probability $\eta$ resampled per sequence. The $\eta{=}0$ component keeps the standard clean-conditioning configuration in-distribution, and 10\% of training batches use the plain pretraining objective (no $K/H$ split) to protect unconditional generation. We instantiate the mixing kernel $q_{\mathrm{noise}}$ two ways: (i) \emph{uniform} over the vocabulary, the classical uniform-mixing kernel of \citet{austin2021structured}; and (ii) a \emph{hard-negative mixture} of equal parts uniform draws, corpus-unigram draws, top-32 embedding-space neighbors of the true token, and masked self-proposals from the base model in the spirit of \citet{clark2020electra}. The latter is a data-dependent kernel concentrated nearer the data conditional. Evaluation-time plausible corruptions always come from a generator family disjoint from all training noise (Section~\ref{sec:setup}), so no arm ever sees its test corruption distribution during training.

\subsection{Objective}

\begin{equation}
\mathcal{L}(\theta) \;=\; \mathbb{E}\!\left[\;\frac{1}{t}\sum_{i \in H} \mathbf{1}\!\left[\tilde{x}^i = \text{M}\right] \mathrm{CE}_i \;+\; \lambda \sum_{i \in K} \mathrm{CE}_i \;\right],
\qquad \mathrm{CE}_i = -\log p_\theta\!\left(x_0^i \mid \tilde{x}\right),
\label{eq:loss}
\end{equation}
with both terms normalized per position. The first term is the standard masked loss over holes; the second teaches copy-else-fix at context positions. We use $\lambda = 0.25$ with uniform position weights, set once and not tuned; the defaults worked on the first attempt, and sweeps over $\lambda$ and substituted-position upweighting are deferred (Section~\ref{sec:limitations}). Equation~\eqref{eq:loss} is a deliberate surrogate rather than the exact variational bound of the composed process: principled ELBOs exist for mixed and interpolating kernels \citep{austin2021structured, campbell2022continuous, vonrutte2025generalized}, and their per-position weighting for the mixing term differs from ours; since $\eta$ is small and the target is editor behavior rather than held-out likelihood, the principled weighting is likewise deferred. We use no auxiliary input flag marking $K$ as potentially corrupted: content-only detection is the more general capability, and hole geometry is already visible through \MASK{}.

\subsection{Inference: reverse-time re-noising of the conditioning}
\label{sec:inference}

Because the finetuned model emits a meaningful distribution at every position, the observed-token probability
\begin{equation}
s_i \;=\; p_\theta\!\left(x_0^i = \tilde{x}^i \mid \cdot\right), \qquad i \in K,
\end{equation}
is a suspicion score for context. The sampler wraps the standard reverse process (Algorithm~\ref{alg:sampler}) with a corrector-like step that acts on the \emph{conditioning} rather than the sample: while $t > t_{\mathrm{lock}} = 0.5$, context tokens with $s_i$ below a threshold annealed from $\tau = 0.5$ to $0.1$ (a time-dependent re-noising intensity) are returned to the absorbing state and join $H$, subject to an edit budget $B = \lceil \hat{\rho}\,|K| \rceil$ with assumed corruption rate $\hat{\rho} = 0.15$; the ordinary reverse dynamics then refill them with full bidirectional conditioning and iterative refinement. Past $t_{\mathrm{lock}}$ the context dynamics freeze. We use 64 reverse steps with greedy commits (with $R$ steps remaining and $m$ masked positions, the $\lceil m/R \rceil$ most confident predictions commit per step). One bookkeeping detail matters more than it looks: context tokens must enter the low-confidence commitment rule of \citet{chang2022maskgit, nie2025llada} with confidence $s_i$, not the indicator value 1, or the sampler never revisits them. A cheap comparator keeps context frozen throughout and applies a single read-out repair at $t = 0$, overwriting positions with $s_i < 0.1$. The re-noising step is a heuristic: it is motivated by predictor--corrector samplers, but we do not derive it from, or show that it preserves, the reverse-time dynamics of the composed forward process.

\begin{figure}[t]
\centering
\fbox{\begin{minipage}{0.94\linewidth}
\textbf{Algorithm 1: annealed detect--remask--refill} \smallskip \hrule \smallskip
\ttfamily\small
r $\leftarrow$ canvas: $H$ fully masked, $K$ = observed context \\
for t = 1 $\to$ 0 in N steps: \\
\hspace*{1.5em} predict $p_\theta(\cdot \mid r)$ at all positions \\
\hspace*{1.5em} if t > t\_lock: \hfill \textrm{\# annealed re-noising of context} \\
\hspace*{3em} for i in K, not yet edited, while edits < B: \\
\hspace*{4.5em} if $s_i < \tau(t)$: r$_i$ $\leftarrow$ \MASK{}; move i into H \\
\hspace*{1.5em} commit / remask hole tokens by low-confidence rule \\
\hspace*{1.5em} \textrm{(context tokens carry confidence $s_i$, not 1)}
\end{minipage}}
\caption{The sampler. Context tokens may be re-noised into the absorbing state only early in the reverse trajectory; repaired positions are refilled by the standard reverse dynamics.}
\label{alg:sampler}
\end{figure}

\section{Experiments}
\label{sec:experiments}

\subsection{Setup}
\label{sec:setup}

\paragraph{Model and training.} We finetune the public 170M-parameter MDLM checkpoint pretrained on OpenWebText \citep{sahoo2024simple, gokaslan2019openwebtext}, GPT-2 tokenizer \citep{radford2019language}, on FineWeb-Edu \citep{penedo2024fineweb} packed into 512-token blocks (400k training blocks, $\approx$205M tokens; evaluation blocks held out from documents 400k positions downstream). Every arm trains for 8{,}000 steps at effective batch $32 \times 512$ ($\approx$131M tokens seen), learning rate $3{\times}10^{-5}$ cosine, bf16, identical seeds. The budget is deliberately modest: under 4.5\,h per arm on a 6\,GB RTX~3060 laptop GPU (21\,min on an RTX~5090). Training dynamics show the intended division of labor: clean-context cross-entropy falls from 3.93 to 0.03 nats (learned trust), substituted-context cross-entropy from 11.2 to 2.5 (learned repair), while hole cross-entropy tracks the vanilla arm (infilling intact).

\paragraph{Evaluation protocol.} Thirteen evaluation cells shared by every arm in a fully paired design (identical hole spans and corruption positions): planted context-corruption rate $\rho \in \{0, 0.02, 0.05, 0.1, 0.2\}$ crossed with $\{$uniform, plausible$\}$ corruptions at 30\% hole coverage, plus coverage $\{10\%, 50\%\}$ at $\rho = 0.1$; 256 passages per cell. Plausible evaluation corruptions are GPT-2 samples conditioned on left context (top-50, true token excluded), a generator family disjoint from every training noise family; plausible-corruption results are therefore out-of-distribution for every arm. Ground truth is the original passage.

\paragraph{Arms.} Table~\ref{tab:arms} lists the nine arms. Arms sharing weights differ only in the sampler, isolating training-time from inference-time contributions; EXT is an external-detector baseline that flags the top-$k$ most surprising context tokens under GPT-2 autoregressive surprisal ($k$ matched per sequence to DP's edit count, which is deliberately generous to the baseline) and refills them with the vanilla arm-A model.

\begin{table}[h]
\caption{Experimental arms. C/D share weights, as do CP/DP.}
\label{tab:arms}
\centering
\small
\begin{tabular}{@{}llll@{}}
\toprule
Arm & Weights & Inference & Isolates \\
\midrule
base & pretrained & frozen context & pre-finetune reference \\
A & vanilla SFT ($\eta = 0$) & frozen context & baseline \\
B & vanilla SFT ($\eta = 0$) & trust-annealed & inference-only contribution \\
C & $\eta$-uniform & frozen $+$ read-out repair & training contribution alone \\
D & $\eta$-uniform & trust-annealed & full method, uniform kernel \\
E & $\eta = 0.15$ fixed, no mixture & trust-annealed & value of the $\eta{=}0$ mixture \\
CP & $\eta$-hard-negative mixture & frozen $+$ read-out repair & data-dependent kernel, cheap inference \\
DP & $\eta$-hard-negative mixture & trust-annealed & full method, data-dependent kernel \\
EXT & vanilla SFT (arm-A refill) & GPT-2 surprisal flags & external-detector baseline \\
\bottomrule
\end{tabular}
\end{table}

\paragraph{Metrics.} Repair: detection precision/recall over planted positions, correction accuracy (edited to the ground-truth token), and the false-edit rate on clean tokens, our primary safety metric. Infill: token-F1 against ground truth, computed per hole span and averaged over all spans in a cell; deltas are reported in points ($\times 100$) with paired-bootstrap 95\% confidence intervals (10k resamples of passages, same statistic as the tables). Passage quality: mean NLL of the completed passage under a gpt2-large judge \citep{radford2019language}, since exact match cannot credit valid-but-different repairs. Containment: infill token accuracy binned by distance to the nearest corrupted context token. Non-regression: clean-context ($\rho = 0$) metrics and the held-out variational bound relative to arm A.

\subsection{Results}
\label{sec:results}

\begin{table}[t]
\caption{Headline cell: plausible corruptions, $\rho = 0.1$, 30\% hole coverage ($n = 256$, paired). Paired-bootstrap F1 deltas vs.\ A (points): DP $+0.51$ \ci{-0.26}{+1.28}; D $+0.38$ \ci{-0.36}{+1.09}; E $+0.51$ \ci{-0.30}{+1.31}; CP $+0.40$ \ci{-0.31}{+1.10}; none is significant.}
\label{tab:main}
\centering
\footnotesize
\begin{tabular}{@{}lcccccc@{}}
\toprule
Arm & Infill F1 & Det.\ P & Det.\ R & False-edit \% & Correction & Judge NLL $\downarrow$ \\
\midrule
base & 0.275 & -- & -- & -- & -- & 3.250 \\
A & 0.287 & -- & -- & -- & -- & 3.237 \\
B & 0.286 & 0.572 & 0.035 & 0.29 & 0.756 & 3.209 \\
C & 0.286 & 0.716 & 0.058 & 0.26 & 0.756 & 3.173 \\
D & 0.291 & 0.668 & 0.150 & 0.83 & 0.730 & 3.083 \\
E & 0.292 & 0.625 & 0.272 & 1.82 & 0.716 & 2.980 \\
CP & 0.291 & \textbf{0.878} & 0.197 & 0.31 & \textbf{0.773} & 3.052 \\
DP & \textbf{0.292} & 0.752 & \textbf{0.350} & 1.28 & 0.743 & \textbf{2.902} \\
EXT & 0.279 & 0.062 & 0.029 & 4.87 & 0.260 & 3.101 \\
\bottomrule
\end{tabular}
\end{table}

\paragraph{Repair (H1).} The training-time forward process appears necessary for the capability: the vanilla model with the same sampler (arm B) detects almost nothing (recall 0.035), because a model whose training dynamics never visit a wrong unmasked state assigns no meaningful likelihood to observed positions; the pretrained checkpoint gives $\approx$50\% probability to uniform-random garbage in context. After the $\eta$-finetune, conspicuous corruptions are caught nearly completely: at uniform $\rho = 0.1$, arm D reaches recall 0.959 \ci{0.954}{0.963} at precision 0.909 with a 1.07\% false-edit rate (DP: recall 0.959 at precision 0.853, 1.84\% false-edit), and corrections land on the exact ground-truth token 52--77\% of the time across $\eta$-arms and corruption types. On out-of-distribution plausible corruptions (Table~\ref{tab:main}), the data-dependent kernel more than doubles recall over the uniform kernel (DP 0.350 vs.\ D 0.150) at precision 0.752, though with a higher false-edit rate (1.28\% vs.\ 0.83\%), consistent with training noise nearer the data conditional transferring better to corruptions drawn near it. Even so, two thirds of plausible corruptions go undetected. False-edit rates on fully clean context stay between 0.07\% and 0.39\% for the mixture-trained arms. At $\rho = 0.2$ recall pins at $\approx$0.76 across $\eta$-arms; this is the $\hat{\rho} = 0.15$ edit budget binding, not a detection ceiling.

\begin{figure}[t]
\centering
\includegraphics[width=0.495\linewidth]{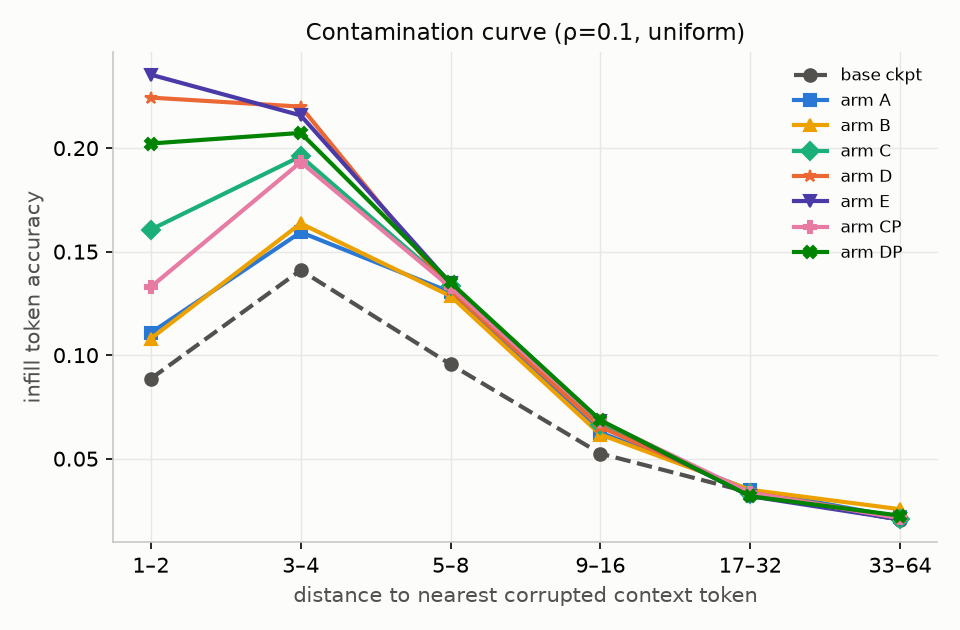}\hfill
\includegraphics[width=0.495\linewidth]{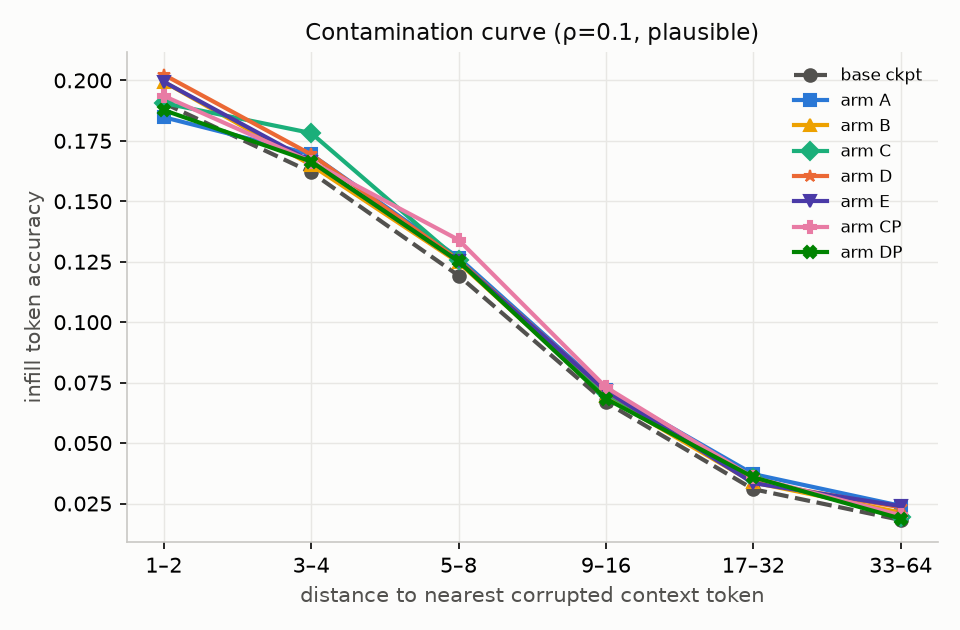}
\caption{Error containment at $\rho = 0.1$: infill token accuracy binned by distance to the nearest corrupted context token. \textbf{Left (uniform):} vanilla arms (base/A/B) dip to 0.088--0.11 adjacent to corruptions, below their own 0.14--0.16 at distance 3--4; $\eta$-arms invert the dip (0.20--0.24 at distance 1--2), and curves converge by distance $\geq 5$. \textbf{Right (plausible):} all arms coincide within noise at every distance; locally plausible substitutions barely contaminate neighboring infills.}
\label{fig:contamination}
\end{figure}

\paragraph{Containment (H2).} Where corruption actually poisons infilling, the method removes the poison. Under uniform corruption at $\rho = 0.1$, vanilla arms show the contamination signature: infill accuracy drops to 0.088--0.11 in the 1--2-token bin adjacent to a corruption, below their own 0.14--0.16 at distance 3--4. Every $\eta$-arm inverts the dip (accuracy adjacent to a now-repaired corruption reaches 0.20--0.24, as high as anywhere in the curve), and all curves converge by distance $\geq 5$: uniform contamination is local, and the method eliminates exactly that local damage (Figure~\ref{fig:contamination}). Under plausible corruption, all arms coincide within noise at every distance. Locally plausible substitutions barely poison neighboring infills even for the vanilla model, so there is little to contain.

\paragraph{Aggregate infill quality.} The pre-registered primary gate, arm D over arm A by $\geq 3$ F1 points at $\rho = 0.1$, passes decisively on uniform corruptions and fails on plausible ones, and we report both unedited. Uniform: D$-$A $= +3.84$ \ci{+2.82}{+4.91} at $\rho = 0.1$ and $+6.42$ \ci{+5.42}{+7.47} at $\rho = 0.2$. At $\rho = 0.1$, E ($+4.47$) and DP ($+3.64$) also clear the gate; C ($+3.15$ \ci{+2.25}{+4.10}) clears it narrowly, and CP ($+2.46$ \ci{+1.67}{+3.29}) improves significantly but falls short. Plausible: D$-$A $= +0.38$ \ci{-0.36}{+1.09}; the best arms, DP and E, reach $+0.51$, small and not statistically significant. At $\rho = 0.2$ plausible, gains are larger and significant (D $+1.19$ \ci{+0.32}{+2.06}, DP $+1.69$ \ci{+0.72}{+2.65}) but still below the gate. We attribute the failure mainly to limited headroom: plausible corruptions cost the vanilla baseline only $\approx$2.3 F1 points at $\rho = 0.1$, so three points of recovery are not available. Low plausible-corruption recall ($\leq 0.35$) also limits the gain, so headroom is not the only factor. The full $13$-cell grid is in Appendix~\ref{app:grid}.

\begin{table}[t]
\caption{Judge-LM passage quality: gpt2-large NLL of the completed passage, nats/token (lower is better), 30\% hole coverage.}
\label{tab:judge}
\centering
\small
\begin{tabular}{@{}lccc@{}}
\toprule
Arm & Clean & Uniform $\rho{=}0.1$ & Plausible $\rho{=}0.1$ \\
\midrule
base & 2.583 & 4.120 & 3.250 \\
A & 2.565 & 4.084 & 3.237 \\
B & 2.578 & 4.014 & 3.209 \\
C & 2.585 & 2.746 & 3.173 \\
D & 2.581 & 2.610 & 3.083 \\
E & 2.593 & 2.626 & 2.980 \\
CP & 2.576 & 2.823 & 3.052 \\
DP & 2.573 & \textbf{2.610} & \textbf{2.902} \\
\bottomrule
\end{tabular}
\end{table}

\paragraph{Passage quality under the judge.} The judge LM measures what exact match cannot (Table~\ref{tab:judge}). Under uniform corruption, repair restores 97\% of the corruption-induced damage: arm A pays $+1.52$ nats/token over its clean reference, arms D/DP pay $+0.03$--$0.04$. Under plausible corruption, DP scores 2.902 vs.\ 3.237 for A ($-0.335$ nats/token; clean reference 2.573). We read this as a passage-level improvement that token-F1 cannot credit, since repairs producing valid-but-different text count as errors under exact match. Judge NLL under a single LM is an indirect proxy, however, and may also reward fluent but unfaithful repairs.

\paragraph{Non-regression of the learned reverse dynamics (H3).} On clean conditioning, paired F1 deltas between treated arms and A straddle zero, judge NLL is identical to within 0.03 nats, and the held-out variational bound moves by $+0.0036$ nats/token (A 3.4072 vs.\ C/D weights 3.4108; gate $\leq 0.02$), with all finetunes improving on the base checkpoint by $\approx$0.35 nats through FineWeb-Edu adaptation. Removing the $\eta{=}0$ mixture (arm E) is the one configuration that pays: persistently elevated clean-context cross-entropy (0.06 vs.\ 0.03 nats), the largest bound delta ($+0.0071$), and the highest false-edit rate at the headline cell (1.82\%). It is a more aggressive editor, at a measurable cost in calibration on clean input. Arm E differs from D in two ways, however: it drops the $\eta{=}0$ component \emph{and} raises the mean context-corruption rate (0.15 vs.\ 0.0375), so the comparison does not isolate the mixture itself; a matched-mean control is future work. We also did not train a spatially homogeneous hybrid-noise arm, so the contrast with reports that such noise pays a downstream cost \citep{vonrutte2025generalized} is indirect.

\paragraph{Annealing (H4).} Moving from frozen-context read-out repair to the annealed sampler on the same weights raises plausible-corruption detection recall from 0.058 to 0.150 (uniform-trained C$\to$D) and from 0.197 to 0.350 (hard-negative-trained CP$\to$DP), at higher false-edit rates (0.26$\to$0.83\% and 0.31$\to$1.28\%). This comparison changes several factors at once: edit timing, the threshold (annealed from 0.5 vs.\ fixed at 0.1), the edit budget, and refilling by iterative reverse steps rather than a single read-out. It therefore supports the sampler as a whole, not annealing specifically; isolating the schedule would require, for example, a fixed-threshold early-edit arm.

\paragraph{External-detector baseline.}
\label{sec:ext}
On uniform corruptions, GPT-2 surprisal flagging is competitive but consistently behind: recall 0.90 vs.\ arm D's 0.96 at $\rho = 0.1$, F1 0.297 vs.\ 0.304, judge NLL 2.701 vs.\ 2.610, with the gap widening at $\rho = 0.2$, where EXT recovers $+4.9$ F1 points over A against arm D's $+6.4$. Against this detector the integrated detect--remask--refill loop is ahead even on conspicuous noise, although EXT receives a per-sequence edit budget matched to DP's and refills with the same class of model. On plausible corruptions the baseline fails structurally, with recall 0.003--0.076 and false-edit rates up to 10.4\%: it flags rare-but-correct tokens, while the corruptions are, by construction, low-surprisal under left context. Its judge NLL of 3.101 lands well short of DP's 2.902 against A's 3.237. We hypothesize that detecting left-context-plausible corruptions requires bidirectional consistency checking, which the $\eta$-finetuned model provides natively (DP: recall 0.350 at 1.28\% false-edit). We tested only this one detector, not bidirectional, larger, or trained ones, so the results do not show that external detectors in general are inferior.

\paragraph{The detectability--damage regularity.} Putting the pieces together: corruptions conspicuous enough to poison neighboring infills (uniform / out-of-distribution) are caught at 96\% recall and their contamination is eliminated; corruptions that evade detection (locally plausible) are precisely those that damage infilling least. We offer a simple stochastic interpretation, which we have not tested quantitatively (we do not measure the divergence directly): both detectability and damage may be governed by how far the corruption kernel departs from the data conditional $p_{\mathrm{data}}(x^i \mid \text{context})$. Plausible corruptions are near-samples from that conditional, and low divergence makes them simultaneously statistically invisible and semantically light; uniform-mixing noise sits far from it, making it at once conspicuous and destructive. Under this reading, the method closes the contamination gap where that divergence is large, and where exact-match gains are capped, the judge suggests that repair still improves passage-level quality. The open question is the corruption class that is near the conditional \emph{and} damaging, entity and number swaps, which our evaluation does not yet cover.

\section{Discussion and limitations}
\label{sec:limitations}

The experiment points to a capability gap created by a collapsed forward-process design space. In our setting the gap narrows substantially for conspicuous corruptions when that space is reopened by composing two classical corruption kernels with region-dependent rates; this is cheap enough to run on a laptop and leaves clean-conditioning behavior intact. It narrows only slightly for plausible ones. The two bracketing failure modes behaved as designed: over-editing is bounded by the inference budget and the false-edit gate (arm E is consistent with the $\eta{=}0$ mixture also helping, subject to the confound noted above), while copy-collapse did not occur (detection recall well above chance in every treated arm). The detectability--damage regularity sharpens what repair is for: not rescuing token-F1 on corruptions too close to the data conditional to matter locally, but eliminating the contamination that divergent corruptions cause, and improving passage-level coherence where a judge can see it.

Several directions follow from the dynamics view: the exact variational bound for region-inhomogeneous composed kernels \citep{campbell2022continuous, vonrutte2025generalized}; per-position rates as a design surface \citep{chen2024diffusion}; principled re-noising policies (threshold, annealing, budget) in place of our fixed heuristics; and combining context trust with ReMDM-style re-noising of committed tokens \citep{wang2025remasking}. The limitations are plain: a single small scale (170M), one seed per arm, and a single text domain (English web text), so generality across sizes and domains is untested; gains under plausible corruption are small; the mixture and annealing ablations are confounded (Section~\ref{sec:results}); the external-detector comparison uses one left-context detector; and the links to reverse-time dynamics and continuous inpainting are motivating analogies, not formal results. Plausible corruptions come from one generator family; entity and number swaps (near-conditional yet damaging), code infilling with injected bugs, pairwise preference evaluation, and the deferred objective ablations are future work. The fixed budget $\hat{\rho} = 0.15$ caps recall at high corruption rates ($\approx$0.76 at $\rho = 0.2$). Scaling to LLaDA-class models \citep{nie2025llada} remains open, as does the conjecture that known-region trust is a single design axis spanning discrete and continuous diffusion.

\bibliographystyle{plainnat}
{\small
\bibliography{refs}
}

\appendix

\section{Full token-F1 grid}
\label{app:grid}

\begin{table}[h]
\caption{Token-F1 for all thirteen shared evaluation cells (256 paired passages each).}
\label{tab:grid}
\centering
\footnotesize
\begin{tabular}{@{}lcccccccc@{}}
\toprule
Cell ($\rho$ / coverage) & base & A & B & C & D & E & CP & DP \\
\midrule
plausible 0.02 / 0.3 & 0.294 & 0.306 & 0.302 & 0.302 & 0.309 & 0.311 & 0.309 & 0.311 \\
plausible 0.05 / 0.3 & 0.284 & 0.304 & 0.299 & 0.304 & 0.305 & 0.310 & 0.304 & 0.306 \\
plausible 0.1 / 0.1 & 0.322 & 0.355 & 0.351 & 0.355 & 0.352 & 0.352 & 0.353 & 0.351 \\
plausible 0.1 / 0.3 & 0.275 & 0.287 & 0.286 & 0.286 & 0.291 & 0.292 & 0.291 & 0.292 \\
plausible 0.1 / 0.5 & 0.252 & 0.259 & 0.259 & 0.254 & 0.258 & 0.261 & 0.264 & 0.263 \\
plausible 0.2 / 0.3 & 0.261 & 0.269 & 0.270 & 0.281 & 0.281 & 0.291 & 0.286 & 0.286 \\
uniform 0 / 0.3 & 0.294 & 0.311 & 0.310 & 0.311 & 0.311 & 0.314 & 0.313 & 0.310 \\
uniform 0.02 / 0.3 & 0.288 & 0.305 & 0.301 & 0.304 & 0.305 & 0.309 & 0.306 & 0.307 \\
uniform 0.05 / 0.3 & 0.279 & 0.291 & 0.290 & 0.305 & 0.305 & 0.313 & 0.312 & 0.306 \\
uniform 0.1 / 0.1 & 0.297 & 0.327 & 0.332 & 0.348 & 0.358 & 0.359 & 0.344 & 0.354 \\
uniform 0.1 / 0.3 & 0.252 & 0.266 & 0.268 & 0.298 & 0.304 & 0.311 & 0.291 & 0.302 \\
uniform 0.1 / 0.5 & 0.217 & 0.225 & 0.225 & 0.252 & 0.262 & 0.269 & 0.255 & 0.262 \\
uniform 0.2 / 0.3 & 0.211 & 0.222 & 0.230 & 0.282 & 0.286 & 0.295 & 0.274 & 0.290 \\
\bottomrule
\end{tabular}
\end{table}

\end{document}